\documentclass[runningheads]{llncs}
\usepackage[T1]{fontenc}
\usepackage{marvosym}
\usepackage{graphicx}
\usepackage{booktabs}
\usepackage{microtype}
\usepackage{amsmath}
\usepackage{amsfonts} 
\usepackage{multirow}
\usepackage{xcolor}
\usepackage{ulem}
\usepackage{enumitem}
\usepackage{tabularx}
\usepackage{makecell}
\usepackage{inconsolata}
\usepackage{graphicx}
\usepackage{tikz}
\usepackage{tikz-qtree}
\usepackage{forest}
\usepackage{graphicx}
\usepackage{algorithmicx}
\usepackage{algorithm}
\usepackage{hyperref}

\begin{document}
\title{SGSimEval: A Comprehensive Multifaceted and Similarity-Enhanced Benchmark for Automatic Survey Generation Systems}
\titlerunning{SGSimEval}
\author{Beichen Guo\inst{1} \and
Zhiyuan Wen\inst{1}$^{\textrm{(\Letter)}}$ \and
Yu Yang\inst{2}$^{\textrm{(\Letter)}}$ \and
Peng Gao\inst{1} \and
Ruosong Yang\inst{3} \and
Jiaxing Shen\inst{4}}

\authorrunning{Guo et al.}
\institute{The Hong Kong Polytechnic University, Hong Kong, China \\
\email{beichen.guo@connect.polyu.hk, zhiyuan.wen@polyu.edu.hk, penggao@polyu.edu.hk} \and
The Education University of Hong Kong, Hong Kong, China \\
\email{yangyy@eduhk.hk} \and
China Mobile (Hong Kong) Innovation and Research Institute, Hong Kong, China \\
\email{yangruosong@cmi.chinamobile.com} \and
Lingnan University, Hong Kong, China \\
\email{jiaxingshen@ln.edu.hk}}
\maketitle % typeset the header of the contribution
\begin{abstract}
The growing interest in automatic survey generation (ASG), a task that traditionally required considerable time and effort, has been spurred by recent advances in large language models (LLMs). With advancements in retrieval-augmented generation (RAG) and the rising popularity of multi-agent systems (MASs), synthesizing academic surveys using LLMs has become a viable approach, thereby elevating the need for robust evaluation methods in this domain. However, existing evaluation methods suffer from several limitations, including biased metrics, a lack of human preference, and an over-reliance on LLMs-as-judges. To address these challenges, we propose SGSimEval, a comprehensive benchmark for \textbf{S}urvey \textbf{G}eneration with \textbf{Sim}ilarity-Enhanced \textbf{Eval}uation that evaluates automatic survey generation systems by integrating assessments of the outline, content, and references, and also combines LLM-based scoring with quantitative metrics to provide a multifaceted evaluation framework. In SGSimEval, we also introduce human preference metrics that emphasize both inherent quality and similarity to humans. Extensive experiments reveal that current ASG systems demonstrate human-comparable superiority in outline generation, while showing significant room for improvement in content and reference generation, and our evaluation metrics maintain strong consistency with human assessments. 
\end{abstract}
\section{Introduction}
\setlength{\parskip}{0pt} 

Academic survey papers synthesize vast literature into coherent narratives, providing crucial scientific overviews. The exponential growth of academic literature has made manual survey creation increasingly time-consuming, driving urgent demands for automation~\cite{10285615,liu2023longtextmultitablesummarization}. Recent advances in Large Language Models (LLMs), Retrieval Augmented Generation (RAG~\cite{gao2023retrieval}), and Multi-Agent Systems (MASs~\cite{han2024llm}) have significantly advanced autonomous literature review~\cite{agarwal2025litllmtoolkitscientificliterature,sami2024systematicliteraturereviewusing,Susnjak_2025,LIU_2022} and scientific discovery~\cite{lu2024aiscientistfullyautomated,weng2025cycleresearcherimprovingautomatedresearch,yuan2025dolphinmovingclosedloopautoresearch}, making automated survey generation (ASG) increasingly feasible.

Recent ASG systems have demonstrated promising capabilities. AutoSurvey~\cite{autosurvey} established foundational methodologies with structured workflows, while SurveyX~\cite{surveyx} enhanced performance through preparation and generation phases. Advanced systems like LLMxMapReduce-V2~\cite{llmmapreducev2} tackle extremely long inputs, SurveyForge~\cite{surveyforge} focuses on outline heuristics, and InteractiveSurvey~\cite{interactivesurvey} introduces personalized generation with user customization.

Despite these advances, ASG evaluation remains problematic. Current frameworks suffer from inconsistent metrics across studies, making cross-system comparisons impossible. While LLM-as-a-Judge approaches~\cite{zheng2023judgingllmasajudgemtbenchchatbot,que2024hellobenchevaluatinglongtext} have gained prominence, existing evaluations inadequately integrate traditional metrics with LLM-based assessments and fail to capture survey quality across structure, references, and content dimensions comprehensively.

To address these limitations, we propose SGSimEval, a comprehensive benchmark for ASG systems through similarity-enhanced assessment. SGSimEval integrates outline quality, content adequacy, and reference appropriateness evaluations while combining LLM-based scoring with quantitative metrics. Our framework introduces two complementary approaches: (1) human-reference alignment and (2) quality-balanced assessment weighing semantic similarity against intrinsic content quality. SGSimEval incorporates 80 highly-cited survey papers from diverse domains, enabling comprehensive assessments through systematic evaluation.

Extensive experiments demonstrate SGSimEval's effectiveness in revealing performance variations across ASG approaches. Results show Computer Science-specialized systems consistently outperform general-domain systems across all dimensions, with most ASG systems surpassing human performance in outline generation and CS-specific systems exceeding humans in content quality, though human-generated references maintain substantially higher quality. Human consistency evaluations through pairwise comparisons confirm that our similarity-enhanced framework provides more nuanced and reliable assessments compared to traditional approaches, with evaluation results aligning well with human preferences across structural organization, content creation, and reference curation dimensions.

This work makes three contributions:
\begin{itemize}[leftmargin=*, topsep=0pt, itemsep=-2pt, partopsep=0pt, parsep=0pt]
    \item We present SGSimEval, a comprehensive benchmark integrating outline quality, content adequacy, and reference appropriateness assessments for ASG evaluation.
    \item We introduce a similarity-enhanced framework with complementary approaches considering both structural alignment and intrinsic content quality, providing better alignment with human preference for survey evaluation.
    \item We provide extensive experimental validation demonstrating our benchmark's effectiveness in identifying meaningful differences between ASG systems.
\end{itemize}
\vspace{-3pt}
\section{Related Works}

\subsection{Automated Survey Generation}
Automated Survey Generation (ASG) systems build upon recent advances in Large Language Models (LLMs), Retrieval Augmented Generation (RAG), and Multi-Agent Systems (MASs) that have accelerated autonomous literature review and scientific discovery~\cite{agarwal2025litllmtoolkitscientificliterature,sami2024systematicliteraturereviewusing,Susnjak_2025,torres2024promptheushumancenteredpipelinestreamline,ali2024automatedliteraturereviewusing,lu2024aiscientistfullyautomated,weng2025cycleresearcherimprovingautomatedresearch,yuan2025dolphinmovingclosedloopautoresearch,schmidgall2025agentlaboratoryusingllm,liu2025visionautoresearchllm}.

Early ASG systems like AutoSurvey~\cite{autosurvey} established foundational, structured workflows. Subsequent research has enhanced these frameworks with more sophisticated techniques, such as decomposing the writing process into distinct phases~\cite{surveyx}, developing strategies for handling extremely long inputs~\cite{llmmapreducev2}, and employing memory-driven generation with learned outline heuristics~\cite{surveyforge}. More recently, the focus has begun to shift towards personalized and interactive systems that allow for continuous user customization~\cite{interactivesurvey}.

\subsection{Evaluation of ASG systems}
The evaluation of generative models has increasingly adopted LLM-as-a-Judge approaches~\cite{zheng2023judgingllmasajudgemtbenchchatbot,que2024hellobenchevaluatinglongtext,wu2025writingbenchcomprehensivebenchmarkgenerative,chen2025mlrbenchevaluatingaiagents}, offering advantages in consistency and efficiency while significantly reducing human annotation costs. In the automated survey generation domain, this trend has been particularly valuable given the complexity and multifaceted nature of survey quality assessment.

Recent advancements in ASG evaluation leverage LLMs alongside traditional metrics and human expert assessments across three core dimensions: structure, references, and content quality. AutoSurvey~\cite{autosurvey} employs a multi-LLM scoring strategy, achieving high consistency with human judgments in citation quality and content relevance, thereby reducing manual annotation costs. Similarly, SurveyForge~\cite{surveyforge} utilizes LLMs to rapidly assess outline rationality, improving structural evaluation efficiency by more than 30\%. These LLM-based approaches are complemented by traditional metrics such as semantic similarity and IoU (reference overlap rate)~\cite{surveyx}, forming a multidimensional evaluation system that ensures model outputs align with both semantic logic and real-world academic scenarios. 

\section{SGSimEval}

\begin{figure}[t]
    \centering
    \includegraphics[width=\textwidth]{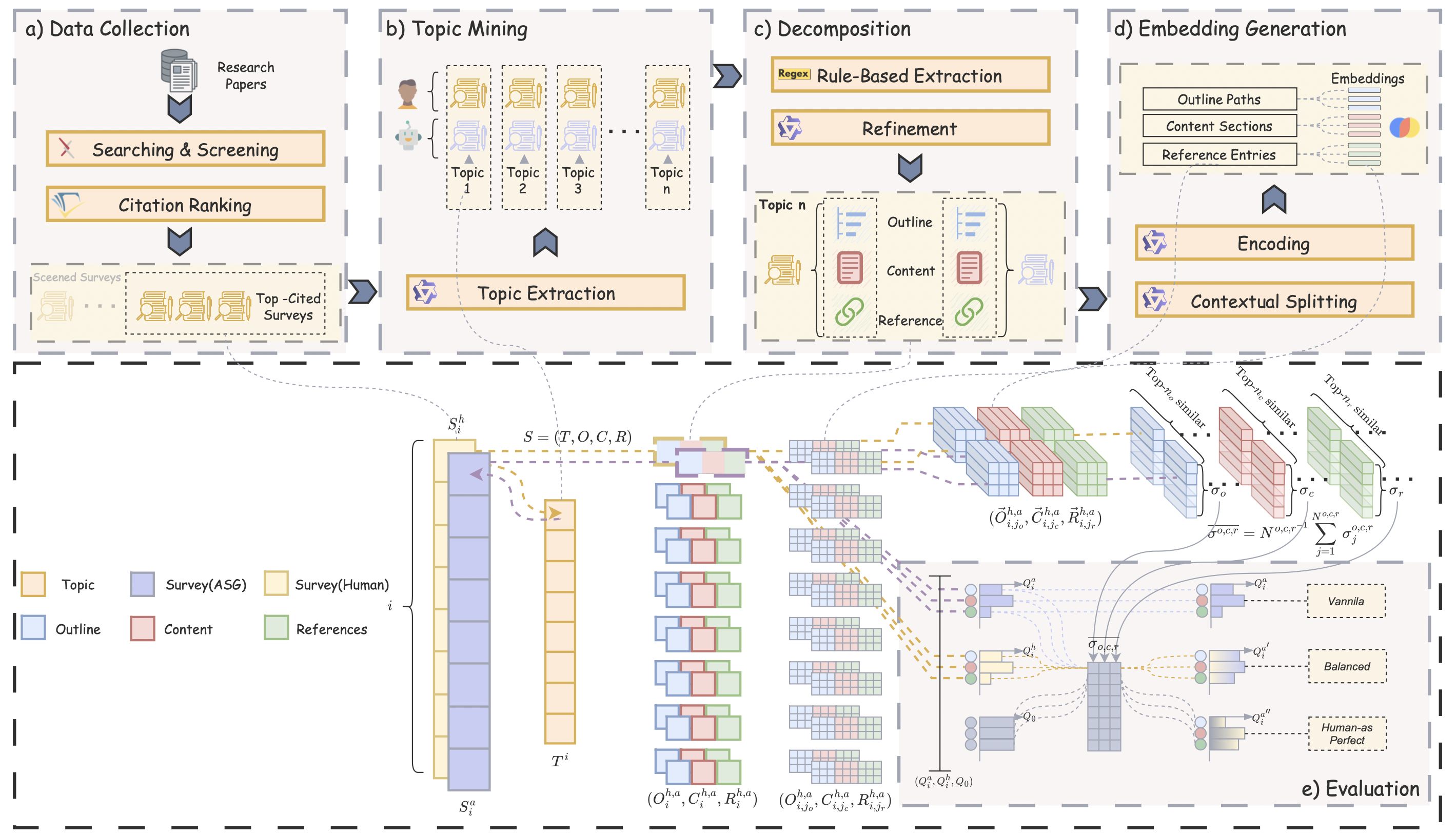}
    \caption{SGSimEval processes all the human-authored surveys and ASG-generated surveys through a) Data Collection, b) Topic Mining, c) Decomposition, d) Embedding Generation, and e) Evaluation. Three types of assessment results are marked as Vanilla, Balanced, and Human-as-Perfect.}
    \label{fig:sgsimeval}
\end{figure}

\begin{figure}[t]
    \centering
    \includegraphics[width=0.4\textwidth]{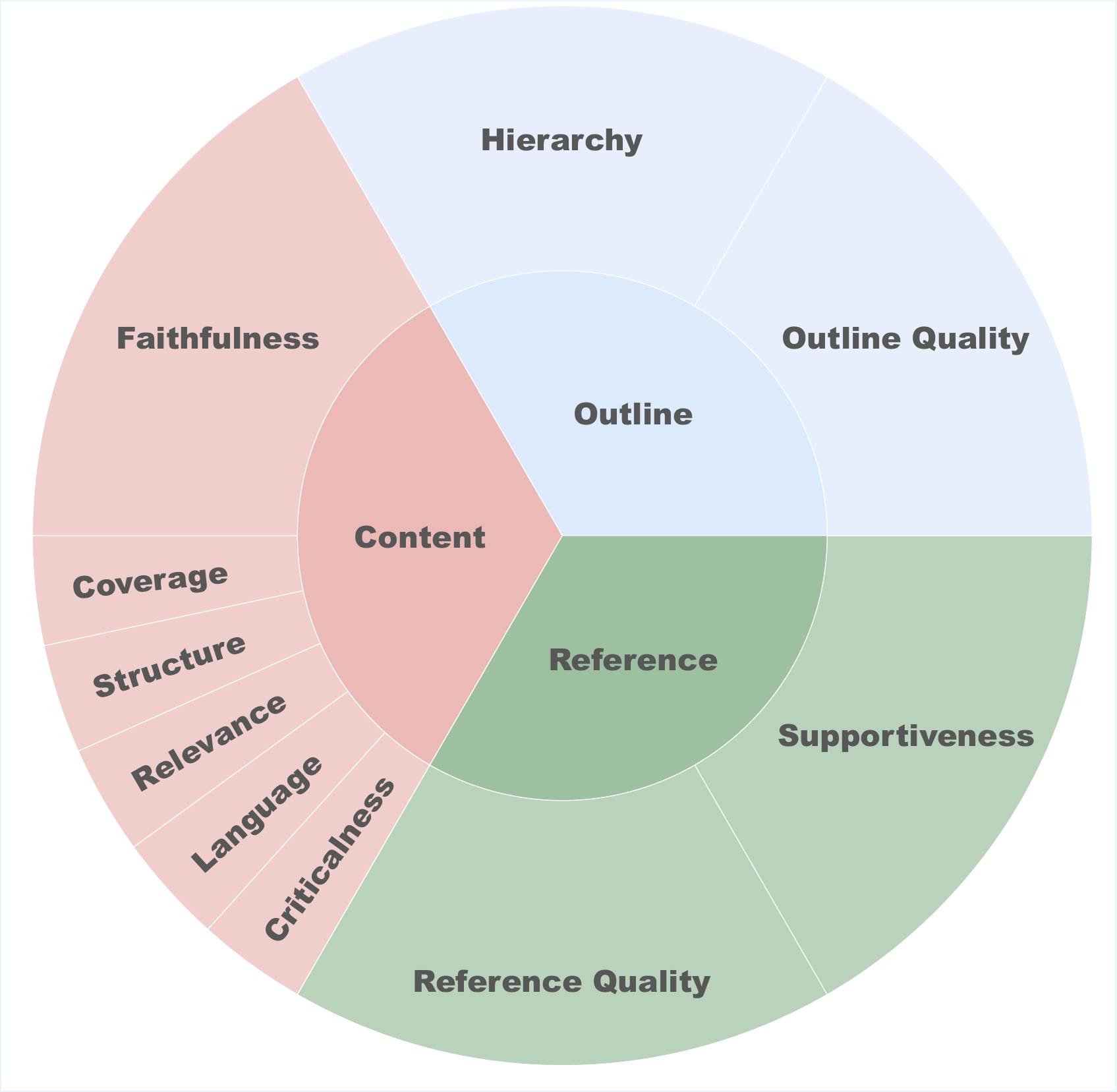}
    \caption{The multifaceted evaluation aspects included in SGSimEval.}
    \label{fig:metrics}
\end{figure}

Figure~\ref{fig:sgsimeval} illustrates the comprehensive pipeline of SGSimEval, which processes human-authored and ASG-generated surveys through five key stages: (a) Data Collection, (b) Topic Mining, (c) Decomposition, (d) Embedding Generation, and (e) Evaluation. Figure~\ref{fig:metrics} provides an overview of the evaluation aspects included in our framework. Our approach begins with the curating of a high-quality dataset of human-authored surveys that undergoes systematic decomposition into outline, content, and references. These components are transformed into contextual embeddings and stored in a vector database for similarity computation. The evaluation stage employs three assessment approaches: Vanilla evaluation using traditional metrics, Human-as-Perfect similarity weighting, treating human content as ideal references, and Balanced similarity weighting, considering both semantic alignment and actual human content quality. This enables a comprehensive assessment of ASG systems across structural, semantic, and quality dimensions while incorporating human preference alignment through automated similarity-based weighting mechanisms.
\subsection{Dataset Curation}
The SGSimEval dataset is curated through a multi-stage process. Initially, $I=80$ highly-cited survey papers are collected and parsed to extract their topics, outlines, content, and references. Subsequently, contextual embeddings $\vec{O}_{i,m_o}$, $\vec{C}_{i,m_c}$, and $\vec{R}_{i,m_r}$ for the extracted components are generated and stored in a vector database for further calculation. Each human-authored survey is represented as $S_i = \{T_i, O_i, C_i, R_i\}$, where $T_i$ denotes topic, $O_i$ denotes outline, $C_i$ denotes content, $R_i$ denotes references, and $i$ ranges from 1 to $I$.
\subsubsection{Data Collection.} 
Our SGSimEval dataset comprises 80 highly-cited literature surveys from arXiv\footnote{\url{https://arxiv.org/}} spanning various domains, all published within the last three years. The surveys are ranked by citation statistics, and we select the top 80 for our dataset. The PDFs of these survey papers undergo initial processing via MinerU~\cite{wang2024mineruopensourcesolutionprecise}, an efficient PDF parsing tool. Subsequently, a rule-based extraction approach facilitates the retrieval of key information, including outlines, content, and references.
\subsubsection{Topic Mining.} 
For all 80 survey papers, we employ LLM to extract high-level topics from their titles. Each title is fed to the model to generate a concise topic label. These labels serve as the primary input for the evaluation of the ASG generation systems (e.g., "Large Language Models Meet NL2Code: A Survey" is labeled as "Natural Language to Code Generation with Large Language Models"). The model's output is then manually verified for accuracy and relevance. This process ensures that the topics are not only relevant but also representative of the content within the survey papers.
\subsubsection{Decomposition.}
We decompose each survey into its constituent components: outline, content, and references. This decomposition process employs rule-based extraction approaches to systematically parse the structured information from each survey paper.
\subsubsection{Contextual Embedding Generation.}
\label{sec:contextual_embedding}
\begin{figure}[t]
    \centering
    \includegraphics[width=0.7\textwidth]{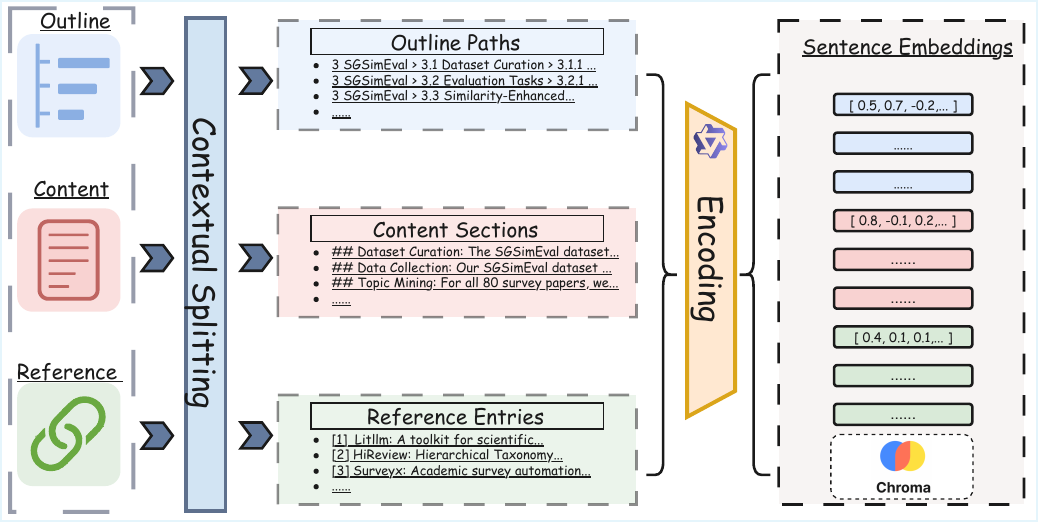}
    \caption{The contextual embedding generation involves splitting the outline, content, and reference into outline paths, content sections, and reference entries, and encoding them into sentence embeddings.}
    \label{fig:split}
\end{figure}
As shown in Figure~\ref{fig:split}, to enable granular similarity comparison, we iteratively split each component type according to specific rules designed to preserve semantic coherence within individual units. For the outlines, a hierarchical section tree is constructed. Each path from the root to a leaf node is extracted. Subsequently, paths corresponding to sibling leaf nodes are aggregated to form distinct documents, for which embedding vectors are then generated. For the textual content, the content of each section is treated as an individual document, and corresponding vector representations are created. Similarly, each bibliographic reference is processed as a discrete document, and its respective embedding vector is generated. All the split embeddings of the outlines, content, and references $\vec{O}_{i,m_o}$, $\vec{C}_{i,m_c}$, and $\vec{R}_{i,m_r}$ are then stored in ChromaDB\footnote{\url{https://www.trychroma.com/}} for further retrieval and comparison.

\subsection{Evaluation Framework}
\subsubsection{SGSimEval-Outline.}
\label{sec:sgeval-outline}
SGSimEval-Outline evaluates the structural quality and logical coherence of generated survey outlines through hierarchical analysis and LLM-based assessment.
\begin{enumerate}
    \item \textbf{Hierarchy:} The hierarchical structure of the generated outlines is evaluated to ensure logical organization and coherence. This is achieved by a depth-based weighted scoring mechanism. For each non-leaf (parent) node $i$ in the outline, a local score $L_i$ is determined. This score $L_i$ represents the proportion of its child nodes that are considered coherent with the parent node, as assessed by LLM. Specifically, the LLM evaluates whether each child logically follows from its parent. Each node $i$ is assigned a weight $w_i$ that is inversely proportional to its depth $d_i$, giving higher importance to nodes closer to the root. The weight is calculated as:
    \begin{equation}
        \label{eq:weight}
        w_i = \frac{D_{\text{max}} - d_i + 1}{D_{\text{max}}}
    \end{equation}
    where $D_{\text{max}}$ is the maximum depth of the outline. The global hierarchy score $H_{\text{outline}}$ is then computed as the weighted average of the local scores of all non-leaf nodes $P$:
    \begin{equation}
        \label{eq:hierarchy_score}
        H_{\text{outline}} = \left( \frac{\sum_{i \in P} L_i \cdot w_i}{\sum_{i \in P} w_i} \right) \times 100
    \end{equation}
    This score reflects the structural integrity of the generated outline.
    \item \textbf{LLM-Assessed Quality:} Following existing studies~\cite{surveyforge}, we also employ the LLM to assess the overall quality of the survey paper outline by LLM-generated criteria.
\end{enumerate}

\subsubsection{SGSimEval-Content.}
\label{sec:sgeval-content}
SGSimEval-Content assesses the quality of generated survey content through citation faithfulness evaluation and comprehensive LLM-based quality assessment across multiple dimensions.
\begin{enumerate}
    \item \textbf{Faithfulness:} The faithfulness of the generated content evaluates how well cited references support the claims made in the text. For each sentence containing citations, we use an LLM to determine whether the cited references support the statement. The faithfulness score $F_{\text{content}}$ is calculated as:
    \begin{equation}
        \label{eq:faithfulness}
        F_{\text{content}} = \frac{\text{Number of Supporting References}}{\text{Total Number of Cited References}} \times 100
    \end{equation}
    This measures how well the content is backed by appropriate citations.
    \item \textbf{LLM-Assessed Quality:} Using LLMs to evaluate survey content based on LLM-generated assessment criteria is commonly employed in existing works~\cite{autosurvey,surveyforge,llmmapreducev2}. However, different studies focus on limited specific aspects. For example,~\cite{autosurvey,surveyx,interactivesurvey} emphasize \textit{Coverage}, \textit{Structure}, and \textit{Relevance} of survey content, while~\cite{llmmapreducev2} concentrate more on \textit{Criticalness} and \textit{Language}. To achieve a more comprehensive assessment, we review prior works and consolidate all assessment metrics. Specifically, in SGSimEval, we evaluate the content quality of survey papers using an LLM across five dimensions as shown in Table~\ref{tab:llm_assessment_dimensions}.
    \begin{table}[th]
    \centering
    \caption{LLM assessment dimensions for content quality evaluation, their sources and descriptions.}
    \label{tab:llm_assessment_dimensions}
    \begin{tabular}{ccc}
    \toprule
    \textbf{Dimension} & \textbf{Source} & \textbf{Description} \\
    \midrule
    \textit{Coverage} & \cite{autosurvey} & Measures how thoroughly the core topics are addressed. \\
    \textit{Structure} & \cite{autosurvey} & Evaluates the logical organization of information. \\
    \textit{Relevance} & \cite{autosurvey} & Considers the pertinence of the content to the survey topic. \\
    \textit{Criticalness} & \cite{llmmapreducev2} & Examines the degree of critical analysis and insight provided. \\
    \textit{Language} & \cite{llmmapreducev2} & Assesses the clarity and professionalism of the writing. \\
    \bottomrule
    \end{tabular}
    \end{table}
\end{enumerate}

\subsubsection{SGSimEval-Reference.}
\label{sec:sgeval-reference}
SGSimEval-Reference evaluates the quality and relevance of generated bibliographies through supportiveness analysis and comprehensive LLM-based reference assessment.
\begin{enumerate}
    \item \textbf{Supportiveness:} The supportiveness of the references evaluates how well the bibliography supports the survey's main topic. For each reference in the generated bibliography, we use an LLM to determine whether it is relevant to the survey topic. The supportiveness score $S_{\text{reference}}$ is calculated as:
    \begin{equation}
        \label{eq:supportiveness}
        S_{\text{reference}} = \frac{\text{Number of Relevant References}}{\text{Total Number of References}} \times 100
    \end{equation}
    This measures how well the bibliography aligns with the survey's subject matter.
    \item \textbf{LLM-Assessed Quality:} To evaluate the overall quality of the references of a survey paper, we also employ LLM for assessment with a specific LLM-generated criterion. The LLM assesses an overall score to the references based on their relevance, diversity, and adequacy in supporting the survey content.
\end{enumerate}
\subsection{Similarity-Enhanced Evaluation Framework}
\label{sec:similarity_enhanced_evaluation}
Human preferences serve as valuable complements to LLM scores~\cite{zheng2023judgingllmasajudgemtbenchchatbot,Bai_2024}. To introduce human preference alignment while automating the evaluation, we compare ASG-generated surveys against a dataset of human-authored surveys. Each survey, whether generated ($S_i^a$) or human-authored ($S_i^h$), consists of an outline ($O_i$), content ($C_i$), and references ($R_i$). To leverage the reference value of human-authored surveys, we propose two similarity-enhanced approaches.

Under the assumption that high-quality academic surveys in established fields often share common structural and semantic patterns, the first approach, \textit{Human-as-Perfect Similarity Weighting}, treats human-authored content as the ideal reference standard. The second, \textit{Balanced Similarity Weighting}, provides a more nuanced evaluation by considering both semantic similarity and the actual quality scores of human-authored content, ensuring that novel, high-quality work is not unfairly penalized for deviating from the human reference.

\subsubsection{Human-as-Perfect Similarity Weighting.}
\label{sec:human_as_perfect}
This scoring strategy emphasizes human-like quality by treating human-authored content as a perfect reference. We assess semantic alignment between system-generated and human-authored components using embedding-based similarity. The embeddings for the outline, content, and references of the generated survey ($\vec{O}_{i}^a, \vec{C}_{i}^a, \vec{R}_{i}^a$) and the human-authored survey ($\vec{O}_{i}^h, \vec{C}_{i}^h, \vec{R}_{i}^h$) are generated via the same pipeline (Section~\ref{sec:contextual_embedding}).

For each component type $k \in \{\text{o, c, r}\}$, we compare each embedding entry from the generated survey, $\vec{V}_{i,j}^{a,k}$, with its most similar counterpart in the human-authored survey, $\vec{V}_{i,j^*}^{h,k}$. We then average the top $N^k$ similarity scores to serve as the weighting factor $\sigma_i^k$.

The weighting factors are calculated as:
\begin{equation}
    \label{eq:similarity_weight}
    \sigma_i^{k} = \frac{1}{N^{k}} \sum_{j=1}^{N^{k}} \cos\left(\vec{V}_{i,j}^{a,k}, \vec{V}_{i,j^*}^{h,k}\right), \quad k \in \{\text{o}, \text{c}, \text{r}\}
\end{equation}
where $\vec{V}^{a,k}$ and $\vec{V}^{h,k}$ are embedding vectors for a component type $k$, $N^{k}$ is the number of top similarity scores considered, and $\vec{V}_{i,j^*}^{h,k}$ are the embeddings of the most similar human-authored counterparts.

The final evaluation score $Q_{i}^{k'}$ combines the similarity score ($\sigma_i^k$) with the system's intrinsic output quality ($Q_{i}^{k}$), assuming human-authored content is perfect with a maximum score of $Q_{0} = 5$:
\begin{equation}
    \label{eq:human_as_perfect}
    Q_{i}^{k'} = \sigma_i^{k} \cdot Q_{0} + (1 - \sigma_i^{k}) \cdot Q_{i}^{k}, \quad k \in \{\text{o}, \text{c}, \text{r}\}
\end{equation}

\subsubsection{Balanced Similarity Weighting.}
This approach provides a more balanced evaluation by considering both semantic similarity and the actual quality scores of human-authored content. The similarity weighting factors, $\sigma_i^k$, are calculated using the same methodology as described above (Equation~\eqref{eq:similarity_weight}). However, this approach incorporates the actual quality scores of the human references.

The final evaluation score $Q_{i}^{k''}$ combines the similarity score with both the human reference quality and the system output quality:
\begin{equation}
    \label{eq:balanced_weighting}
    Q_{i}^{k''} = \sigma_i^k \cdot Q_{\text{human},i}^k + (1 - \sigma_i^k) \cdot Q_i^k, \quad k \in \{\text{o}, \text{c}, \text{r}\}
\end{equation}
where $Q_{\text{human}, i}^k$ is the quality score obtained by processing the human-authored surveys through the same evaluation pipeline. This formulation provides a more realistic assessment by considering the actual quality variations in human-authored content.

\section{Experiments}
\subsection{Experiment Settings}
For all the LLM-as-Judge scoring and NLI tasks, we use \texttt{Qwen-Plus-2025-04-28}\footnote{\url{https://www.aliyun.com/product/bailian}\label{fn:note1}} with a temperature of 0.5 and a context window of 128k. To ensure full reproducibility, the complete prompts used for these evaluations are provided alongside our source code on GitHubl\footnote{\url{https://github.com/TechnicolorGUO/SGSimEval}} due to space constraints. For all vector models, we use Alibaba's \texttt{Text-Embedding-V3}\footref{fn:note1} with 1024 dimensions.
As detailed in Section~\ref{sec:similarity_enhanced_evaluation}, for the similarity scores, we consider $N^o=N^c=5$ for outline, content and $N^r = 20$ for reference. These values were chosen to balance precision and recall; a smaller $N$ for outline and content focuses on the most critical structural and semantic similarities, while a larger $N$ for references allows for a more comprehensive comparison of the cited literature.

\subsection{Evaluated ASG Systems}

To compare existing Automatic Survey Generation (ASG) systems, we evaluate outputs from five representative tools: InteractiveSurvey~\cite{interactivesurvey}, SurveyX~\cite{surveyx}, AutoSurvey~\cite{autosurvey}, SurveyForge~\cite{surveyforge}, and LLMxMapReduce-V2~\cite{llmmapreducev2}, using the same topic set as our benchmark. Due to open-source and deployment constraints, AutoSurvey and SurveyForge are restricted to the \textit{Computer Science} domain. We achieved an overall successful collection rate of 85.38\% across all systems.

\section{Results}
\subsection{Performance Comparison across Systems}
\vspace{5pt}
\begin{table}[th]
    \centering
    \caption{Performance of different systems using SGSimEval across outline, content, and reference metrics. Results are shown for vanilla, SGSimEval-B (Balanced Weighting), and SGSimEval-HP (Human as Perfect weighting) configurations. All scores are normalized to a 0-5 scale (quantitive scores with 0-100 scale are marked with "*"). Subscript values indicate the original percentage scores before normalization. For Outline and Reference, L1 represents overall quality. For Content, L1-L5 correspond to Coverage, Structure, Relevance, Language, and Criticalness, respectively.}
    \label{tab:experiment_results}
    \centering
    \scriptsize
    \setlength{\tabcolsep}{2pt}
    \renewcommand{\arraystretch}{1.2}
    \resizebox{\textwidth}{!}{
    \begin{tabular}{l
        |l
        |c
        |c c
        |c c c c c c
        |c c 
        }
    \toprule
    \multirow{2}{*}{\textbf{System}} 
    & \multirow{2}{*}{\textbf{Metric}}
    & \multirow{2}{*}{\textbf{Avg.}}
    & \multicolumn{2}{c|}{\textbf{Outline}} 
    & \multicolumn{6}{c|}{\textbf{Content}} 
    & \multicolumn{2}{c}{\textbf{Reference}} \\
    \cmidrule(lr){4-13}
    & 
    & 
    & \makecell{L1}
    & \makecell{*Hierarchy} 
    
    & \makecell{L1} 
    & \makecell{L2} 
    & \makecell{L3} 
    & \makecell{L4} 
    & \makecell{L5} 
    & \makecell{*Faithfulness}
    
    & \makecell{L1} 
    & \makecell{*Supportiveness}\\
    
    \midrule
    \multicolumn{13}{l}{\underline{Domain-Specific Systems (Computer Science)}}   \\
    \midrule
    \textbf{AutoSurvey} & \textit{Vanilla} & 3.85 & 4.00 & $4.68_{93.57}$ & 5.00 & 4.60 & 5.00 & 5.00 & 4.90 & $3.49_{69.75}$ & 3.70 & $3.39_{67.74}$ \\
    & \textit{SGSimEval-B} & 4.00 & 4.02 & $4.77_{95.31}$ & 5.00 & 4.27 & 5.00 & 4.61 & 4.83 & $4.08_{81.58}$ & 4.03 & $3.14_{62.89}$ \\
    & \textit{SGSimEval-HP} & 4.54 & 4.65 & $4.89_{97.70}$ & 5.00 & 4.86 & 5.00 & 5.00 & 4.96 & $4.46_{89.17}$ & 4.41 & $4.28_{85.55}$ \\
    \midrule
    \textbf{SurveyForge} & \textit{Vanilla} & 4.33 & 4.00 & $4.88_{97.50}$ & 5.00 & 4.00 & 5.00 & 5.00 & 4.60 & $4.04_{80.79}$ & 4.60 & $4.07_{81.34}$ \\
    & \textit{SGSimEval-B} & 4.17 & 4.00 & $4.83_{96.67}$ & 5.00 & 4.07 & 5.00 & 4.60 & 4.73 & $4.28_{85.65}$ & 4.41 & $3.39_{67.75}$ \\
    & \textit{SGSimEval-HP} & 4.75 & 4.59 & $4.95_{98.97}$ & 5.00 & 4.67 & 5.00 & 5.00 & 4.87 & $4.67_{93.37}$ & 4.84 & $4.62_{92.37}$ \\
    \midrule
    \multicolumn{13}{l}{\underline{General Domain Systems}}   \\    
    \midrule
    \textbf{InteractiveSurvey} & \textit{Vanilla} & 3.31 & 3.60 & $4.06_{81.28}$ & 4.70 & 3.94 & 4.86 & 4.87 & 4.04 & $3.59_{71.88}$ & 3.23 & $2.28_{45.64}$ \\
    & \textit{SGSimEval-B} & 3.49 & 3.69 & $4.34_{86.82}$ & 4.79 & 4.13 & 4.94 & 4.65 & 4.30 & $3.79_{75.80}$ & 3.45 & $2.34_{46.71}$ \\
    & \textit{SGSimEval-HP} & 4.00 & 4.29 & $4.57_{91.31}$ & 4.87 & 4.56 & 4.94 & 4.95 & 4.59 & $4.42_{88.31}$ & 3.66 & $3.01_{60.29}$ \\
    \midrule
    \textbf{LLMxMapReduce-V2} & \textit{Vanilla} & 3.28 & 4.02 & $4.88_{97.63}$ & 4.86 & 4.02 & 5.00 & 4.40 & 4.33 & $2.23_{44.59}$ & 2.32 & $2.71_{54.30}$ \\
    & \textit{SGSimEval-B} & 3.56 & 3.94 & $4.71_{94.26}$ & 4.84 & 4.14 & 4.99 & 4.43 & 4.42 & $3.25_{65.00}$ & 3.15 & $2.71_{54.27}$ \\
    & \textit{SGSimEval-HP} & 4.19 & 4.58 & $4.95_{98.98}$ & 4.94 & 4.64 & 5.00 & 4.77 & 4.75 & $4.00_{79.92}$ & 3.41 & $3.64_{72.77}$ \\
    \midrule
    \textbf{SurveyX} & \textit{Vanilla} & 3.50 & 3.78 & $4.84_{96.74}$ & 4.06 & 3.35 & 5.00 & 3.97 & 3.61 & $3.70_{73.92}$ & 3.01 & $1.95_{39.04}$ \\
    & \textit{SGSimEval-B} & 3.61 & 3.83 & $4.69_{93.87}$ & 4.51 & 3.84 & 4.99 & 4.24 & 4.09 & $3.77_{75.43}$ & 3.70 & $2.36_{47.18}$ \\
    & \textit{SGSimEval-HP} & 4.31 & 4.49 & $4.93_{98.63}$ & 4.59 & 4.28 & 5.00 & 4.55 & 4.38 & $4.45_{89.04}$ & 4.05 & $3.54_{70.88}$ \\
    \midrule
    \multicolumn{13}{l}{\underline{Human Written}}   \\
    \midrule
    \textbf{Human} & \textit{Vanilla} & 3.74 & 3.89 & $4.56_{91.10}$ & 4.84 & 4.21 & 4.99 & 4.45 & 4.49 & $3.82_{76.42}$ & 4.30 & $2.84_{56.70}$ \\
    % & \textit{SGSimEval} & 3.74 & 3.89 & $4.56_{91.10}$ & 4.84 & 4.21 & 4.99 & 4.45 & 4.49 & $3.82_{76.42}$ & 4.30 & $2.84_{56.70}$ \\
    % & \textit{SGSimEval-HP} & 3.74 & 3.89 & $4.56_{91.10}$ & 4.84 & 4.21 & 4.99 & 4.45 & 4.49 & $3.82_{76.42}$ & 4.30 & $2.84_{56.70}$ \\
    \bottomrule
    \end{tabular}
    }
    \label{tab:experiment_results}
\end{table}
\vspace{-10pt} 

\begin{figure}[t]
    \centering
    \includegraphics[width=1\textwidth]{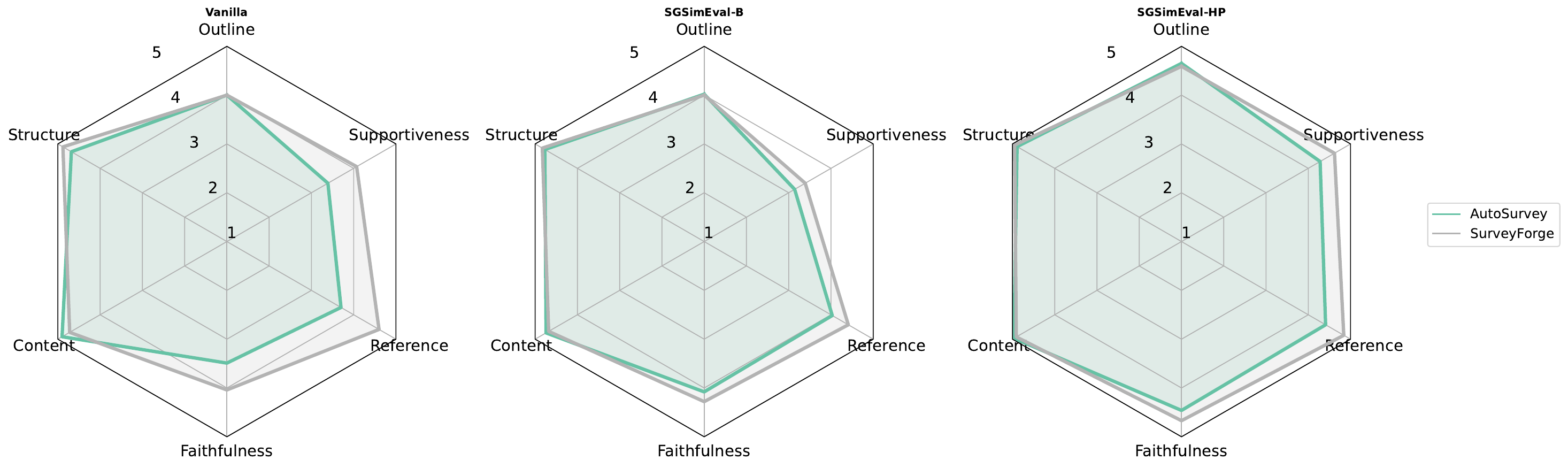}
    \caption{Performance comparison for Computer Science domain systems on vanilla, SGSimEval-B, and SGSimEval-HP configurations: SGSimEval-HP (Human-as-Perfect) strongly rewards alignment with the human reference, while SGSimEval-B (Balanced) provides a more nuanced assessment by balancing similarity with intrinsic quality.}
    \label{fig:experiment_results_cs}
\end{figure}

\begin{figure}[t]
    \centering
    \includegraphics[width=1\textwidth]{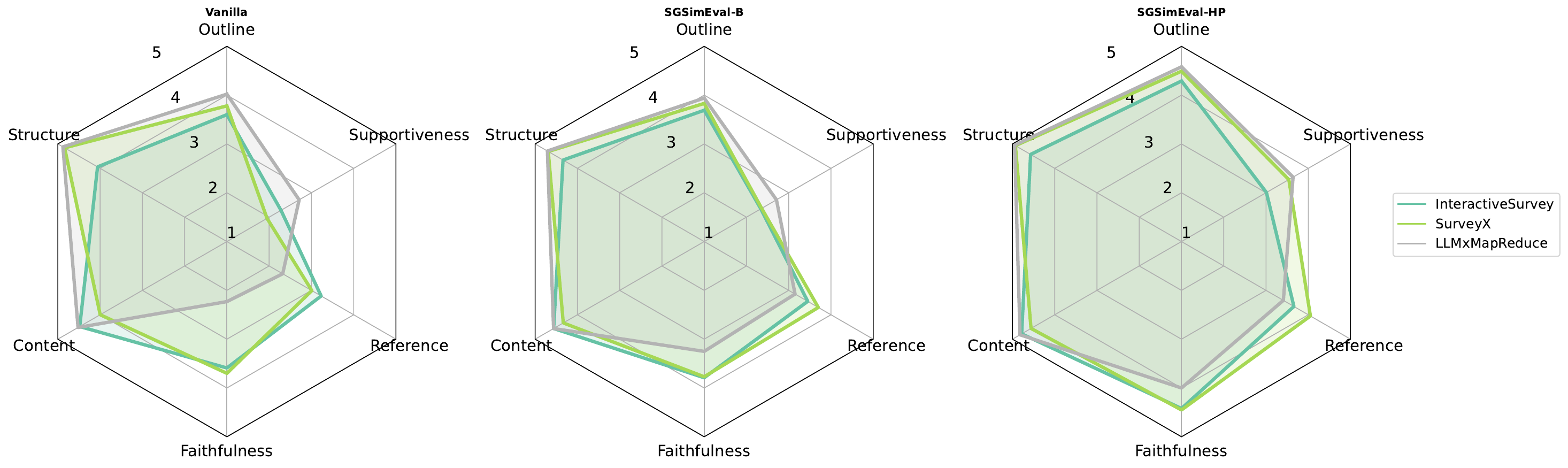}
    \caption{Performance comparison for general domain systems on vanilla, SGSimEval-B, and SGSimEval-HP configurations.}
    \label{fig:experiment_results_general}
\end{figure}

We present the performance difference among the evaluated ASG systems in Table~\ref{tab:experiment_results}, Figures~\ref{fig:experiment_results_cs} and \ref{fig:experiment_results_general}.

CS-specific systems (AutoSurvey and SurveyForge~\cite{autosurvey,surveyforge}) consistently outperform general-domain systems across all dimensions, a superiority likely stemming from their use of domain-specific databases. Within their respective categories, certain systems show distinct strengths: SurveyForge~\cite{surveyforge} excels in outline structure and reference quality; LLMxMapReduce-V2~\cite{llmmapreducev2} leads in outline generation; and SurveyX~\cite{surveyx} demonstrates superior content quality. However, systems reliant on general online search tend to exhibit weaker reference quality. 

\subsection{Performance Comparison between Human and ASG Systems}
Then, we also identify significant differences between ASG systems and human-generated surveys across evaluation dimensions, as shown in Table~\ref{tab:experiment_results}.

When compared with human-authored surveys, ASG systems show a mixed performance profile. Most systems match or exceed human performance in generating structurally coherent outlines, though achieving human-level holistic quality remains competitive. For content quality, domain-specific systems~\cite{autosurvey,surveyforge} outperform humans, while general-domain systems~\cite{interactivesurvey,surveyx,llmmapreducev2} achieve comparable results. However, humans maintain a substantial lead in reference quality. This is further evidenced by the similarity scores in Table~\ref{tab:experiment_results_similarity}, where reference similarity (0.32-0.61) shows the greatest variation and lags significantly behind content similarity (0.57-0.67), confirming that reference selection remains the most challenging aspect for current ASG systems.

\begin{table}[t]
    \caption{Results of the similarity comparison between ASG Systems and human-written surveys}
    \label{tab:experiment_results_similarity}
    \centering{
    \begin{tabular}{c|c c c c}
    \toprule
    \textbf{System} & \textbf{Outline} & \textbf{Content} & \textbf{Reference} & \textbf{Avg.}\\
    \midrule
    \textbf{AutoSurvey} & \textbf{0.63} & 0.65 & 0.56 & 0.61 \\
    \textbf{SurveyForge} & 0.59 & \textbf{0.67} & \textbf{0.61} & \textbf{0.62} \\
    \textbf{InteractiveSurvey} & 0.54 & 0.58 & 0.32 & 0.48 \\
    \textbf{LLMxMapReduce-V2} & 0.58 & 0.64 & 0.42 & 0.54 \\
    \textbf{SurveyX} & 0.60 & 0.57 & 0.56 & 0.58 \\
    \hline
    \textbf{Human} & 1.00 & 1.00 & 1.00 & 1.00 \\
    \bottomrule
    \end{tabular}
    }
    \label{tab:experiment_results_similarity}
\end{table}

\subsection{Human Consistency}

To assess consistency between human preference and our evaluation framework, we conduct LLM-based pairwise comparisons following~\cite{idahl2025openreviewerspecializedlargelanguage}. Using Qwen2.5-72B-Instruct~\cite{qwen2025qwen25technicalreport}, Qwen-Plus-2025-04-28, and Qwen-Turbo-2025-04-28~\cite{yang2025qwen3technicalreport}, we compute average win rates for each ASG system against human baselines across outline, content, and reference dimensions, as shown in Table~\ref{tab:experiment_consistency}.

The pairwise comparison results align with our quantitative metrics. ASG systems achieve higher win rates in outline generation, confirming their effectiveness in structural organization. However, win rates for content quality and reference selection indicate that ASG systems remain competitive but require improvement to consistently match human performance. This highlights the ongoing challenge of achieving human-level proficiency in nuanced content creation and reference curation, where domain expertise and critical judgment remain essential.

\begin{table}[t]
    \caption{ASG systems' Win Rate against human baselines.}
    \label{tab:experiment_consistency}
    \centering
    \begin{tabular}{l|l|c|c}
        \toprule
        \textbf{System} & \textbf{Outline} & \textbf{Content} & \textbf{Reference} \\
        \midrule
        \textbf{AutoSurvey} & $0.73_{\pm 0.15}$ & $0.30_{\pm 0.17}$ & $0.10_{\pm 0.10}$ \\
        \midrule
        \textbf{SurveyForge} & $0.63_{\pm 0.15}$ & $0.33_{\pm 0.35}$ & $\mathbf{0.33_{\pm 0.32}}$ \\
        \midrule
        \textbf{InteractiveSurvey} & $0.44_{\pm 0.11}$ & $0.37_{\pm 0.15}$ & $0.07_{\pm 0.06}$ \\
        \midrule
        \textbf{LLMxMapReduce-V2} & $\mathbf{0.73_{\pm 0.05}}$ & $\mathbf{0.45_{\pm 0.09}}$ & $0.02_{\pm 0.00}$ \\
        \midrule
        \textbf{SurveyX} & $0.63_{\pm 0.12}$ & $0.30_{\pm 0.21}$ & $0.14_{\pm 0.06}$ \\
        \bottomrule
    \end{tabular}
\end{table}

% \begin{table}
%     \centering 
%     \scriptsize
%     \setlength{\tabcolsep}{2pt}
%     \renewcommand{\arraystretch}{1.2}{
%     \begin{tabular}{l|l|c|c}
%     \hline
%     \textbf{System} & \textbf{Outline} & \textbf{Content} & \textbf{Reference} \\
%     \hline
%     \textbf{AutoSurvey} & $0.73_{\pm 0.15}$ & $0.30_{\pm 0.17}$ & $0.10_{\pm 0.10}$ \\
%     \hline
%     \textbf{InteractiveSurvey} & $0.44_{\pm 0.11}$ & $0.37_{\pm 0.15}$ & $0.07_{\pm 0.06}$ \\
%     \hline
%     \textbf{LLMxMapReduce} & $0.73_{\pm 0.05}$ & $0.45_{\pm 0.09}$ & $0.02_{\pm 0.00}$ \\
%     \hline
%     \textbf{SurveyForge} & $0.63_{\pm 0.15}$ & $0.33_{\pm 0.35}$ & $0.33_{\pm 0.32}$ \\
%     \hline
%     \textbf{SurveyX} & $0.63_{\pm 0.12}$ & $0.30_{\pm 0.21}$ & $0.14_{\pm 0.06}$ \\
%     \hline
%     \end{tabular}
%     }
%     \caption{Simplified Evaluation Metrics for Survey Generation Systems Across Different Models.}
%     \label{tab:human_consistency}
% \end{table}

\section{Limitations}
We acknowledge the inherent limitations of the "LLM-as-a-Judge" paradigm. LLMs may exhibit biases (e.g., towards certain writing styles), creating a potential for circular reasoning, as our framework evaluates human-authored surveys using the same pipeline as generated ones. We mitigate this by focusing on relative performance, ensuring any systematic bias is applied uniformly for a fair comparison across systems. However, objective validation remains a challenge. While we employed a state-of-the-art model and fixed prompting to ensure consistency, future work must include direct comparisons with human expert scores to robustly validate the reliability of our LLM evaluator.

\section{Conclusion}
In this paper, we propose SGSimEval, a comprehensive benchmark for evaluating automated survey generation systems with three key contributions: (1) a multidimensional evaluation framework that systematically assesses structural organization, content adequacy, and reference appropriateness; (2) a similarity-enhanced evaluation approach that balances human-authored references with semantic alignment; and (3) extensive experimental validation across five representative ASG systems. Our results demonstrate that CS-specialized systems consistently outperform general-domain approaches, with most ASG systems exceeding human performance in outline generation, though significant challenges remain in reference curation.

\section*{Acknowledgement}
This work was conducted at the Research Institute for Artificial Intelligence of Things (RIAIoT) and supported by PolyU Internal Research Fund (No.BDZ3) and PolyU External Research Fund (No.ZDH5). Also, this work has benefited from the financial support of the EdUHK project under Grant No. RG 67/2024-2025R and Lingnan University (SDS24A5).

% \clearpage
% \appendix
% \section{Evaluation Criteria}
% \label{sec:appendix-criteria}
% \input{tables/criteria.tex}
\bibliographystyle{splncs04}
\bibliography{mybibliography}




% LNCS Homepage, \url{http://www.springer.com/lncs}, last accessed 2023/10/25
% \end{thebibliography}
\end{document}